%% file: main.tex
\documentclass[twocolumn]{article}

\usepackage{framed,multirow}

\usepackage{amssymb}
\usepackage{latexsym}
\usepackage{graphicx}
\usepackage{natbib}
\usepackage{url}
\usepackage{xcolor}
\usepackage{amsmath}
\usepackage{textcomp}
\usepackage{etoolbox}
\usepackage{fancyhdr}
\usepackage{authblk}
\definecolor{newcolor}{rgb}{.8,.349,.1}

\begin{document}

\title{SceneAdapt: Scene-Based Domain Adaptation for Semantic Segmentation using Adversarial Learning}
\date{}
\author[1,2]{Daniele Di Mauro}  
\author[1]{Antonino Furnari} 
\author[2]{Giuseppe Patan\`e} 
\author[1]{Sebastiano Battiato} 
\author[1]{Giovanni Maria Farinella}
\affil[1]{Department of Mathematics and Computer Science, University of Catania, Italy}
\affil[2]{Park Smart s.r.l., Catania, Italy}
\maketitle

\begin{abstract}
Semantic segmentation methods have achieved outstanding performance thanks to deep learning. 
Nevertheless, when such algorithms are deployed to new contexts not seen during training, it is necessary to collect and label scene-specific data in order to adapt them to the new domain using fine-tuning.
This process is required whenever an already installed camera is moved or a new camera is introduced in a camera network due to the different scene layouts induced by the different viewpoints.
To limit the amount of additional training data to be collected, it would be ideal to train a semantic segmentation method using labeled data already available and only unlabeled data coming from the new camera. 
We formalize this problem as a domain adaptation task and introduce a novel dataset of urban scenes with the related semantic labels.
As a first approach to address this challenging task, we propose SceneAdapt, a method for scene adaptation of semantic segmentation algorithms based on adversarial learning.
Experiments and comparisons with state-of-the-art approaches to domain adaptation highlight that promising performance can be achieved using adversarial learning both when the two scenes have different but points of view, and when they comprise images of completely different scenes.
\\To encourage research on this topic, we made our code available at our web page: \url{https://iplab.dmi.unict.it/ParkSmartSceneAdaptation/}.
\end{abstract}


\input{intro.tex}

\input{related.tex}

\input{datasets.tex}

\input{method.tex}
\input{results.tex}

\section{Conclusion}
\label{sec:conclusions}
We have investigated the problem of scene-based domain adaptation for semantic segmentation by defining 
two adaptation scenarios: point of view adaptation and scene adaptation. 
To this aim, we collected a dataset of urban scenes and proposed a novel architecture to perform scene and point of view semantic segmentation adaptation using adversarial learning. 
Experiments show that the proposed method greatly reduces 
over-fitting in both point of view and scene adaptation and outperforms baselines and 
other state-of-the-art methods. Future work can be devoted to combining explicit geometric priors (as in the case of the WARP baseline), with the adversarial framework adopted by the proposed method. 
in the case of point of view adaptation.
\bibliographystyle{apalike}
\bibliography{bibliography}

\end{document}

%% file: intro.tex
\section{Introduction}
\label{sec:intro}
Semantic segmentation allows to classify each pixel of an image according to the object it belongs to in the scene.
This task is often the first step in scene understanding and is important in many industrial scenarios including video surveillance and traffic analysis~\citep{raymond2001traffic,Battiato2018208,ravi2016semantic}.
In recent years, deep learning allowed to push forward the performance of semantic segmentation methods~\citep{badrinarayanan2015segnet2,long2015fully,zhao2017pyramid,7913730}.
However, such approaches generally require to be trained on large quantities of domain-specific labeled data in order to achieve reasonable performance.
In practice, deploying a semantic segmentation system often involves collecting domain-specific data and fine-tuning an existing semantic segmentation algorithm pre-trained on large quantities of domain-agnostic data.
Moreover, this fine-tuning process (including the collection and labeling of new data) needs to be repeated whenever the target domain changes significantly, e.g., when an existing camera is moved or a new one is introduced in the camera network, due to the different scene layouts induced by the different viewpoints.
The significant effort required to collect and label domain-specific data can slow down the deployment of industrial systems based on semantic segmentation algorithms and prevent them from scaling up.

\begin{figure}[t!]
	\centering
	\includegraphics[width=1\linewidth]{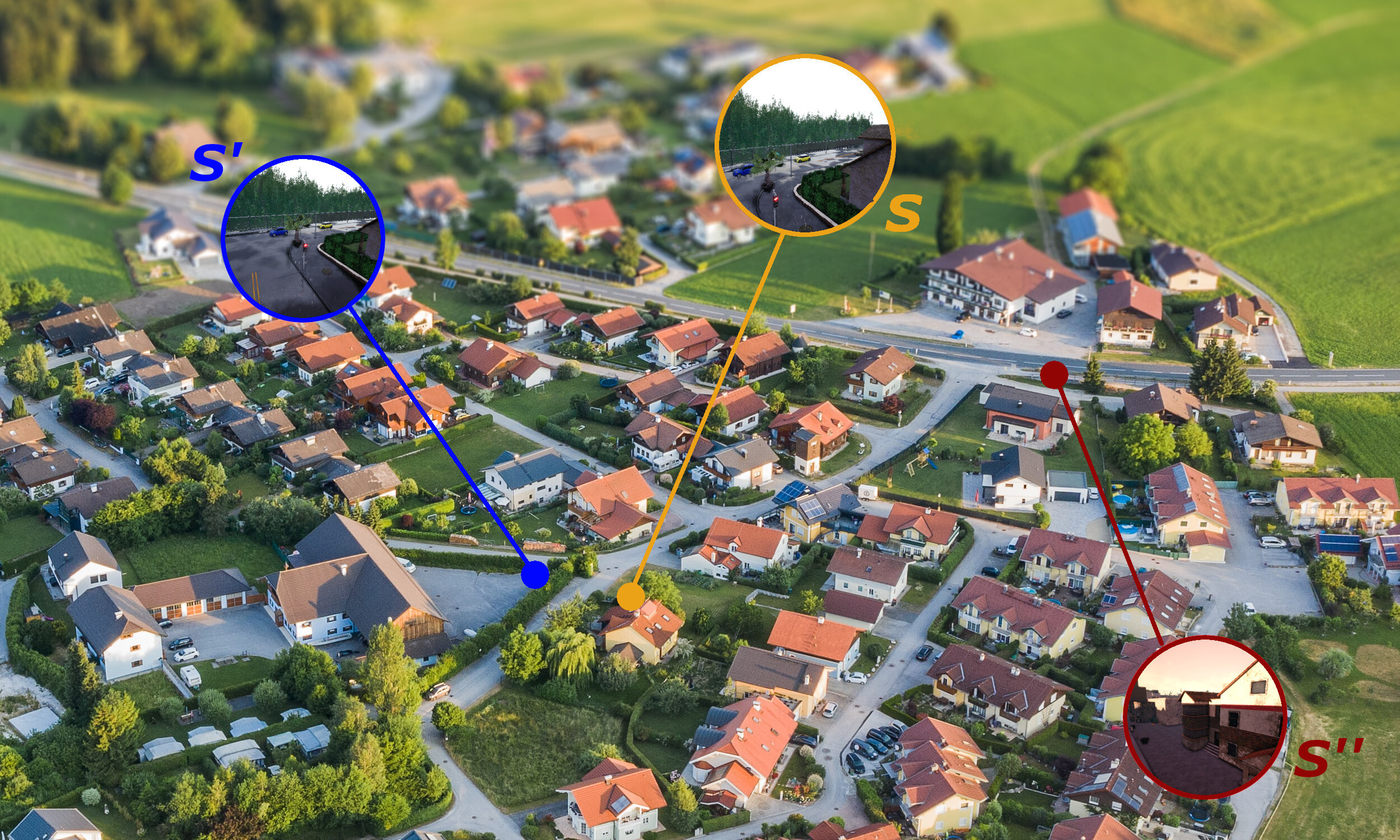}
	\caption[]{The proposed scene adaptation scenario\footnotemark. A camera $S$ (dark yellow) is installed to monitor a urban space. This initial set up requires the collection and labeling of 
		domain-specific data to optimize an existing semantic segmentation algorithm to the considered task. The system can be extended by adding a camera which 
		observes the same scene from a different point of view $S'$ (blue) or by adding a new camera which observes a different scene $S''$ (red). In both cases, the existing semantic segmentation algorithms 
		need to be adapted to the new views.}
	\label{fig:scene_adaptation}
\end{figure}

In this paper, we consider a scene adaptation scenario in which a fixed camera is employed to monitor a urban area. When the system is set up, domain-specific data is collected and labeled. 
An existing algorithm is hence fine-tuned to perform the semantic segmentation of the scene. 
In real scenarios, the system is often extended by adding a new camera looking at the same scene from a different point of view or by adding a camera looking at a different scene characterized by the same semantic classes (e.g. a different urban area). 
After the installation of the new cameras, the existing algorithms need to be adapted to the new views or scenes. 
The proposed scene adaptation problem is illustrated in
\figurename~\ref{fig:scene_adaptation}. 
The adaptation procedure in such contexts is generally achieved collecting and labeling additional data from the two cameras in order to fine-tune the algorithms in a purely supervised way. 
However, as observed in previous work~\citep{DBLP:journals/corr/TzengHSD17,DBLP:journals/corr/abs-1802-10349,DBLP:journals/corr/abs-1711-06969}, the amount of effort required to collect and label domain-specific data to perform fine-tuning can be reduced using domain adaptation techniques (i.e., trying to bypass the three steps of data collection, labeling and fine-tuning).

\footnotetext{Underneath photo by Felix Steininger on Unsplash.}
We frame the considered scene adaptation task as a Domain Adaptation problem in which labeled images collected using the already installed camera belong to the source domain, whereas unlabeled images collected using the new camera belong to the target domain.
We observe that no datasets comprising real images from different cameras looking at the same/similar scenes are available to support our investigation.
This is mainly due to the challenges arising when collecting real data in these settings.
To overcome these issues and provide a first set of data useful to investigate the proposed problem, we collect and publicly release a novel dataset of urban scenes using the CARLA Urban Driving Simulator~\citep{Dosovitskiy17}.
The collected dataset comprises images generated in $3$ scene contexts from $6$ different views ($2$ views per context).

As a first step towards addressing the proposed task, we contribute a method which leverages Adversarial Learning~\citep{goodfellow2014generative} to perform scene adaptation. 
The proposed approach takes advantage only of existing \textit{labeled} data from the source domain and newly collected \textit{unlabeled} images from the target domain. 
We enforce that the input images can be reconstructed from the produced segmentation maps, while the realness of the reconstructions is imposed through adversarial learning.
The reconstructive criterion allows to define a supervised loss for the unlabeled data and acts as a regularizer which encourages the network to learn view invariant representations for small scene elements such as traffic signs and fences.
While the proposed approach is conceptually simple, we show that it significantly outperforms a pipeline including the geometric warp of the input images, which suggests that it is appropriate to consider the problem as a domain adaptation task. We hope that the proposed approach may serve as a useful baseline for other works.
The method is independent from the specific semantic segmentation network employed and can be regarded to as a general training procedure which allows generalization to unseen views.
Experiments show that the proposed approach achieves state of the art performance, surpassing baselines and previous domain adaptation techniques.

The contributions of this work are as follows: 1) we introduce the problem of scene-based domain adaptation for semantic segmentation algorithms; 2) we propose CARLA-SA (Carla-Scene Adaption), a dataset of urban scenes to study the considered domain adaptation problem; 3) we contribute SceneAdapt, a method to perform scene adaptation using adversarial learning, which is shown to outperform baseline and state-of-the-art approaches.
This work extends our previous investigation~\citep{dimauro18} introducing a clearer introduction of the proposed method, additional comparisons with state of the art approaches and baselines, as well as an ablation study discussing the impact of each component on the final performance. 
Moreover, with this paper we publicly release both the collected dataset and the code of the proposed method\footnote{\url{http://iplab.dmi.unict.it/SceneAdapt/}.}.

%% file: related.tex
\section{Related Work}
\label{sec:related}
\textbf{Semantic Segmentation\hspace{2mm}} Many recent works have proposed semantic segmentation algorithms based on deep learning. Among the most notable recent approaches, \cite{long2015fully} proposed fully convolutional networks, which are obtained by adapting existing CNNs for classification to produce dense image segmentation masks. 
\cite{badrinarayanan2015segnet2} presented SegNet, a semantic segmentation approach based on an encoder-decoder architecture.
\cite{7913730} combined artrous convolution and artrous spatial pyramid pooling (ASPP) to segment objects at multiple scales, thus capturing both objects and image context. 
\cite{DBLP:journals/corr/ZhaoQSSJ17} proposed a cascade network to achieve real-time semantic segmentation. 
\cite{zhao2017pyramid} introduced PSPNet, which comprises a novel pyramid pooling module to leverage global contextual information. 
The scene-adaptation approach presented in this paper can integrate any semantic segmentation method which is trainable end-to-end. In our experiments, we will consider the state-of-the-art PSPNet architecture.

\textbf{Adversarial learning\hspace{2mm}} The proposed approach leverages the paradigm of adversarial learning, originally introduced with Generative Adversarial Networks (GANs) by~\cite{goodfellow2014generative}. 
In particular, our work is related to the investigations of~\cite{isola2017image,CycleGAN2017}, who addressed the problem of image translation using GANs. The image translation problem assumes the presence of two different domains $X$ and $Y$. The goal of image translation is to learn a function $F : X \rightarrow Y$ which maps an input $x \in X$ to an output $\hat{y}$ which is indistinguishable from the elements sampled from $Y$. 

\textbf{Domain Adaptation for Semantic Segmentation\hspace{2mm}} The use of domain adaptation techniques for semantic segmentation has also been investigated in past works. \cite{hoffman2016fcns}, proposed a semantic segmentation network to perform global domain alignment with fully convolutional domain adversarial learning. 
In a subsequent work,~\cite{hoffman2017cycada} proposed to adapt representations at the pixel-level and at the feature-level with cycle-consistency without requiring aligned pairs. This model has been applied to different visual recognition and prediction settings, including the semantic segmentation of road scenes. \cite{DBLP:journals/corr/abs-1711-06969} designed a domain adaptation method for semantic segmentation composed by an embedding network, a pixel-wise classifier, a generator network and a discriminator network.
\cite{DBLP:journals/corr/abs-1802-10349} employed a multi-level adversarial network to perform output space domain adaptation at different feature levels. The algorithm consists of two modules: a segmentation network and a discriminator operating at different levels of the network.
Some works address the domain gap created by the device used to capture the data~\cite{yang2019pass}
for instance addressed the domain gap between pinhole and omnidirectional perspectives, whereas~\cite{xiang2020boosting}
addressed semantic segmentation based on a dual-camera system with different perspectives and overlapped views. \cite{romera2019bridging}, 
focused on the domain gap due to the moment of the day in which a given a scene has been captured. 

Our method leverages some of the intuitions presented in these prior works, framing the investigation in the context of the proposed scene-adaptation task.

%% file: datasets.tex
\begin{figure}[th!]
\hspace{0.02\linewidth}
	\begin{minipage}{0.7\linewidth}
		\scriptsize
		\begin{tabular}{cc|c}
			& \multicolumn{2}{c}{\includegraphics[width=1\textwidth]{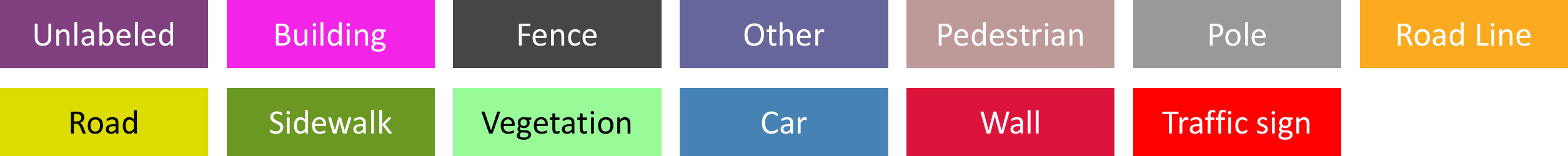}} \\
			
			\raisebox{4\normalbaselineskip}[0pt][0pt]{\rotatebox[origin=c]{90}{\textbf{$A1$}}} &
			\includegraphics[width=0.49\textwidth]{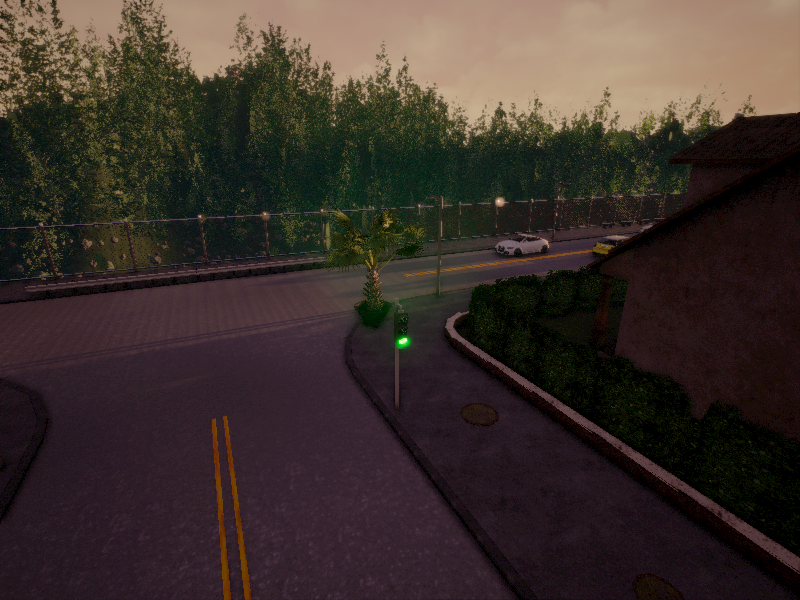} &
			
			\includegraphics[width=0.49\textwidth]{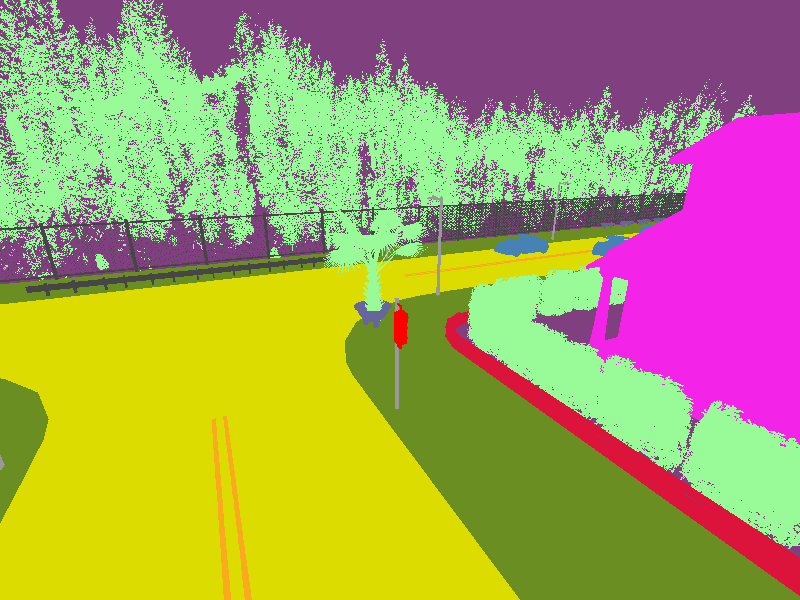} \\
			
			\raisebox{4\normalbaselineskip}[0pt][0pt]{\rotatebox[origin=c]{90}{\textbf{$B1$}}} &
			\includegraphics[width=0.49\textwidth]{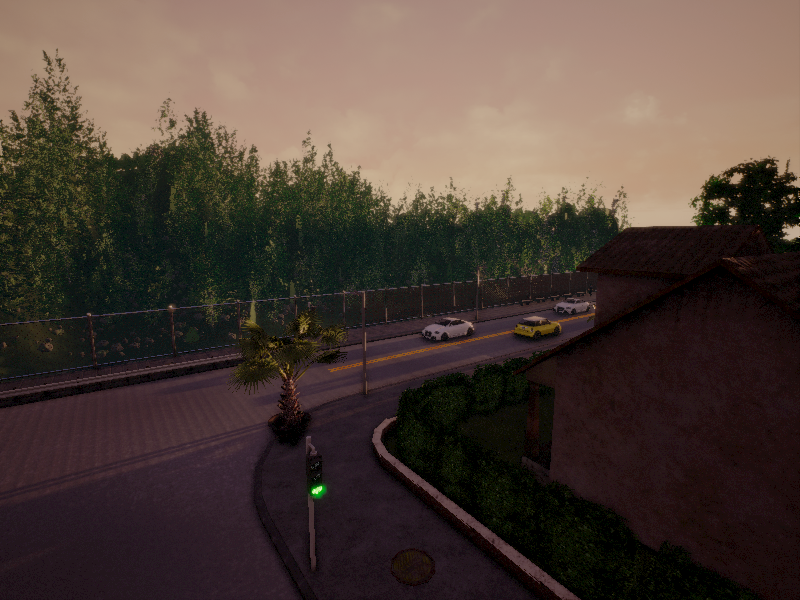} &
			
			\includegraphics[width=0.49\textwidth]{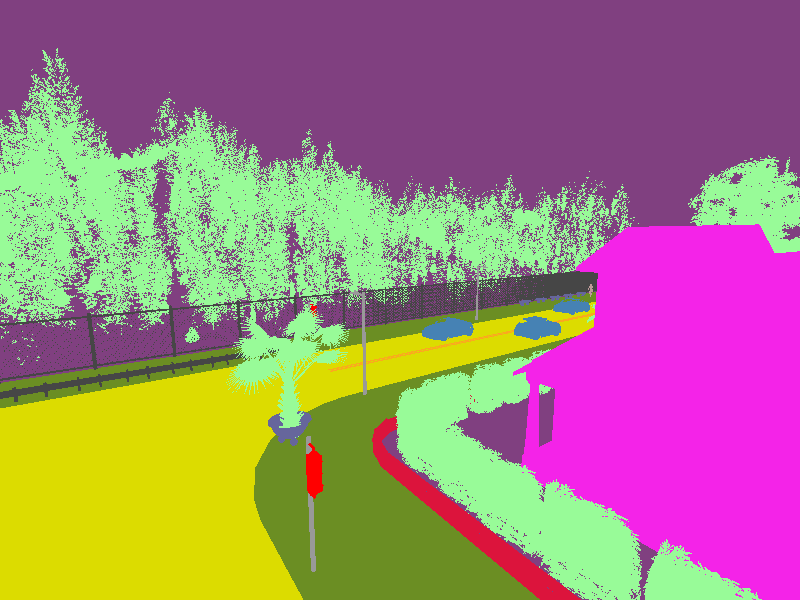}  \\
			
			\raisebox{4\normalbaselineskip}[0pt][0pt]{\rotatebox[origin=c]{90}{\textbf{$A2$}}} &
			\includegraphics[width=0.49\textwidth]{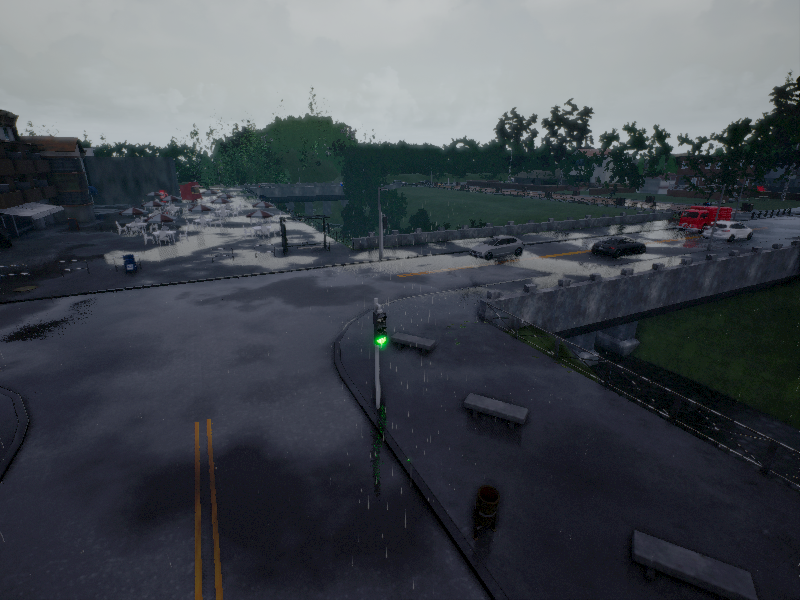} &
			
			\includegraphics[width=0.49\textwidth]{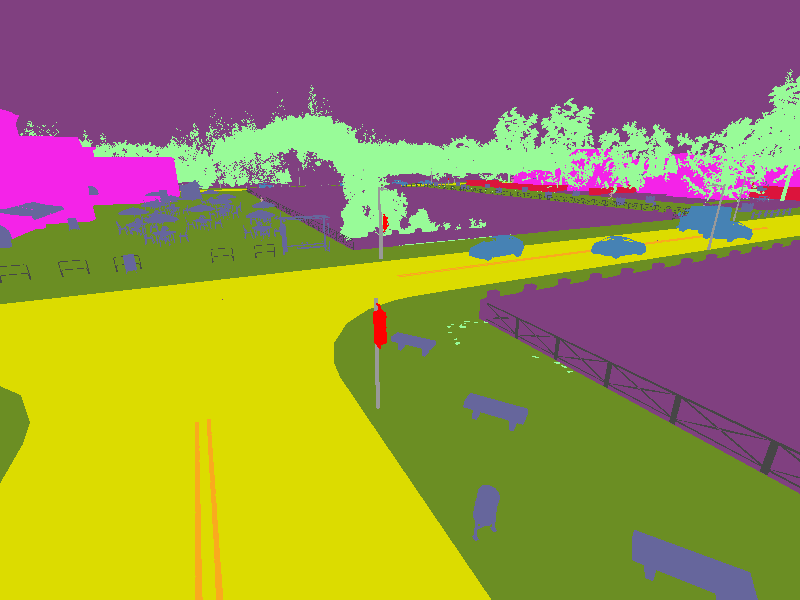} \\
			
			\raisebox{4\normalbaselineskip}[0pt][0pt]{\rotatebox[origin=c]{90}{\textbf{$B2$}}} &
			\includegraphics[width=0.49\textwidth]{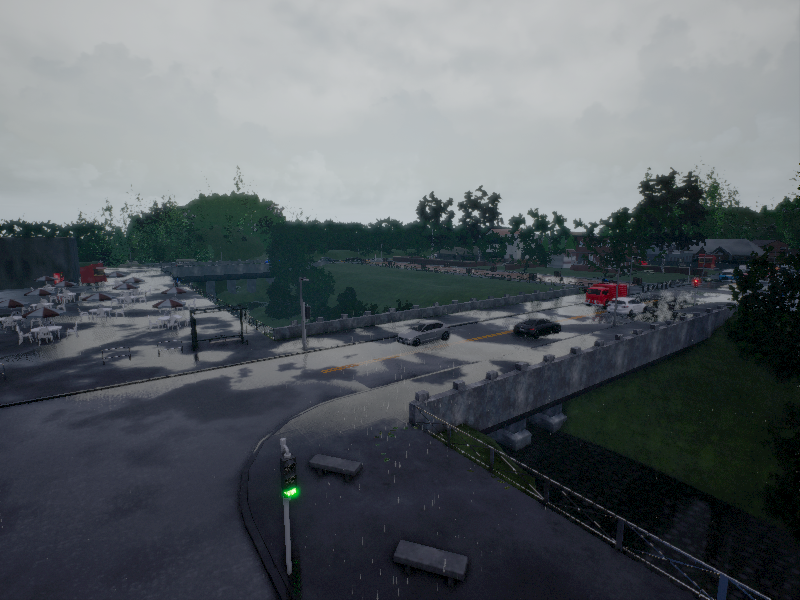}  &
			
			\includegraphics[width=0.49\textwidth]{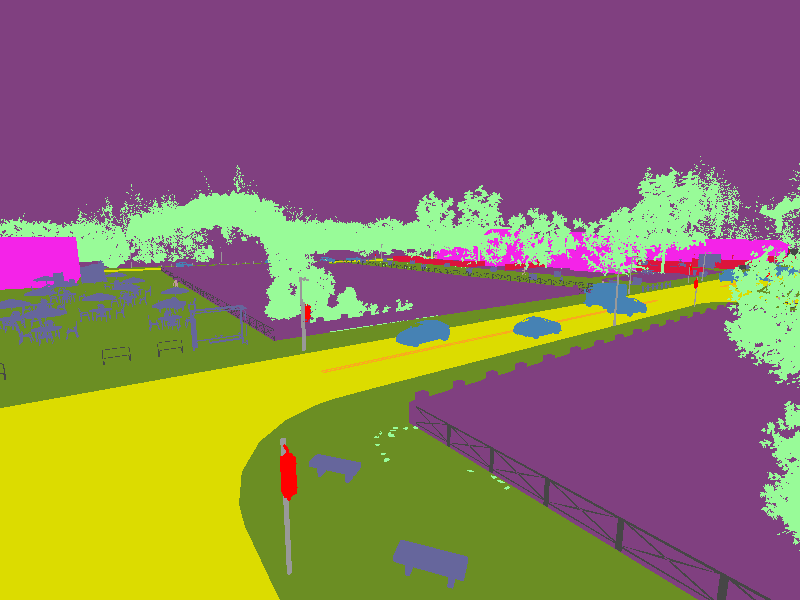} \\
			
			\raisebox{4\normalbaselineskip}[0pt][0pt]{\rotatebox[origin=c]{90}{\textbf{$A3$}}} &
			\includegraphics[width=0.49\textwidth]{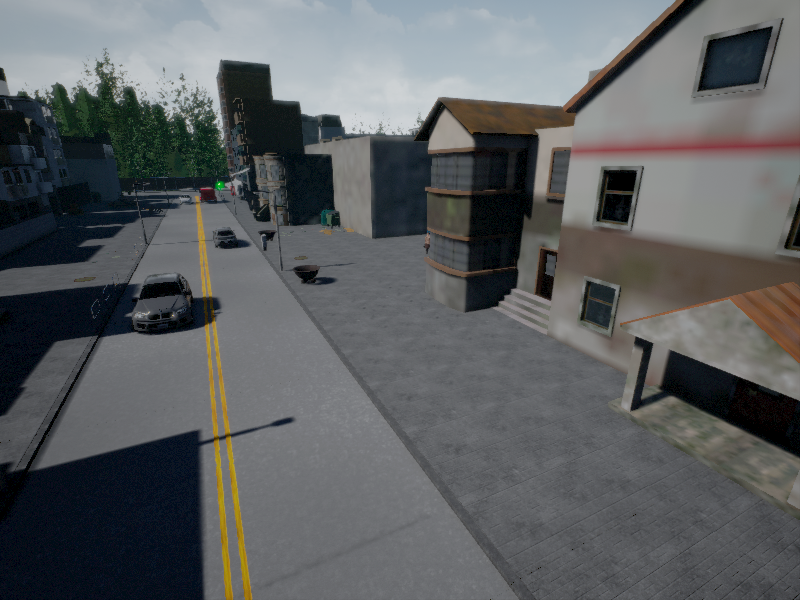}  &
			
			\includegraphics[width=0.49\textwidth]{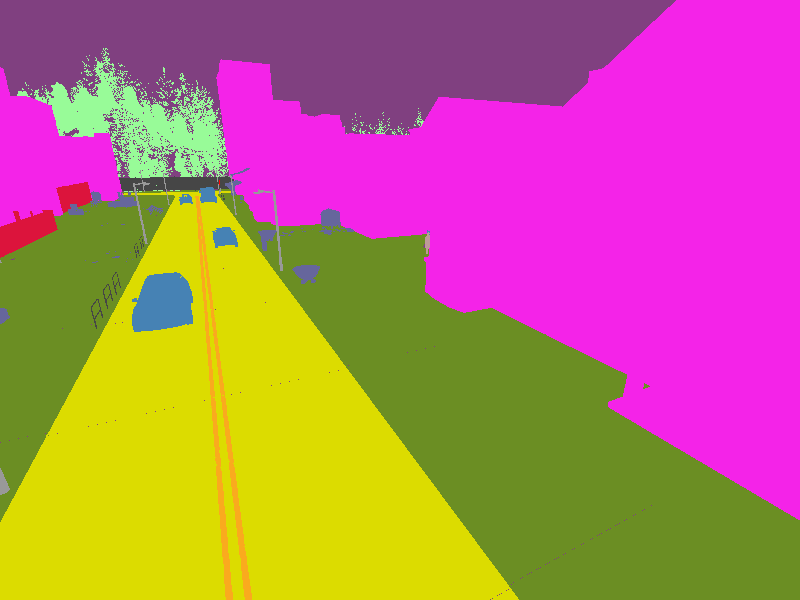} \\
			
            \raisebox{4\normalbaselineskip}[0pt][0pt]{\rotatebox[origin=c]{90}{\textbf{$B3$}}} &
			\includegraphics[width=0.49\textwidth]{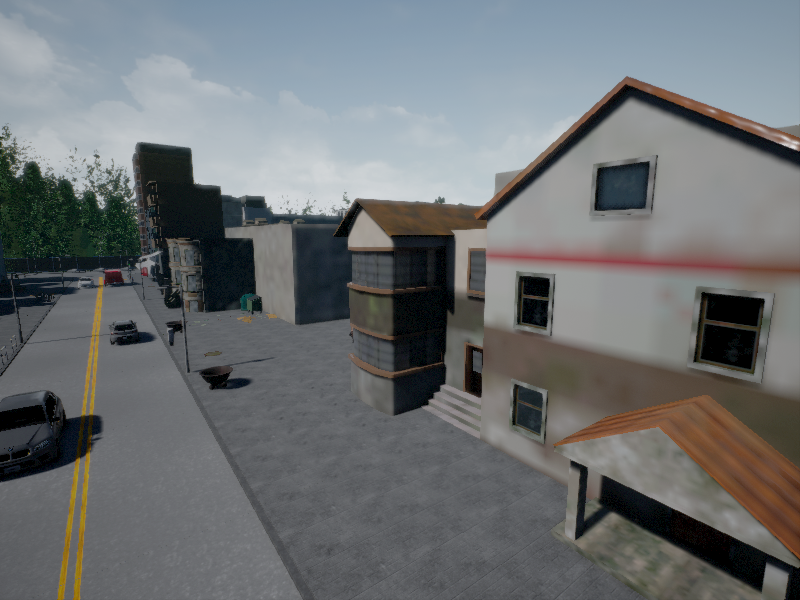}  &
			
			\includegraphics[width=0.49\textwidth]{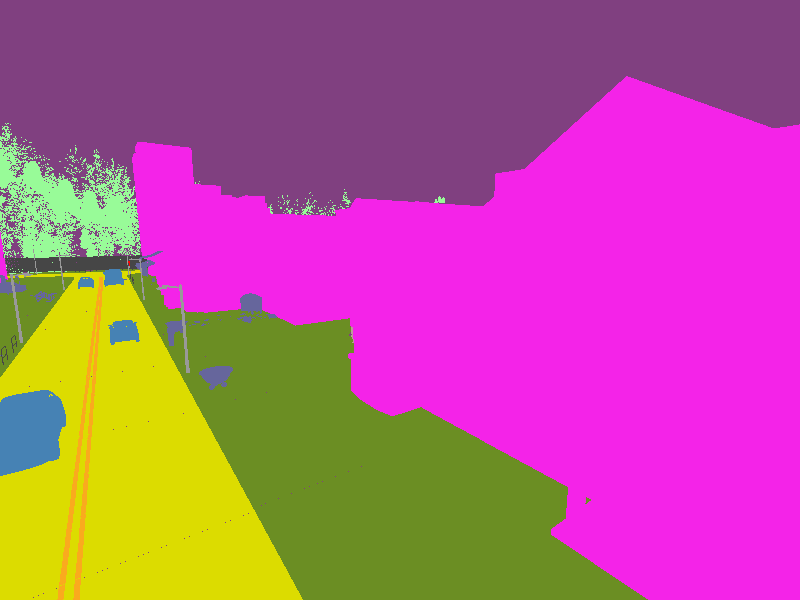} \\

		\end{tabular}
	\end{minipage}
	\caption{Sample images and ground truth segmentation masks from the $6$ data subsets.}
    \label{fig:dataset}
\end{figure}

\section{Dataset}
\label{sec:datasets}
We collected the dataset CARLA-SA using the CARLA Urban Driving Simulator~\citep{Dosovitskiy17}. The simulator has been designed to generate images from the point of view of vehicles for autonomous driving, whereas, to study the scene adaptation problem, we need to collect images of the same scene from different fixed points of view. 
We achieve this by placing the vehicle which collects the images at predefined locations in the virtual city. 
To obtain different views of the same scene, for each vehicle we set up multiple cameras at different altitudes, pitch, yaw and roll angles. The vehicle collecting the images does not move during the simulations, while other vehicles and pedestrians are free to move in the scene.

Using this procedure, we collected three episodes in three different scene contexts of the virtual city, which we refer to as ``Scene 1'', ``Scene 2'' and ``Scene 3''. 
Each scene has been captured by two different points of view presenting scene overlap, which we refer to as ``View $A$'' and ``View $B$''. The two views differ by an angle of 10\textdegree of pitch and by an angle of 10\textdegree of yaw. Therefore, the dataset consists of $6$ image subsets denoted as ``XY'', where $X$ represent the view and $Y$ represents the scene characterizing the subset (e.g., $A1$, $B2$, etc.). \figurename~\ref{fig:dataset} reports some examples from the $6$ subsets. Each subset is split into a training and a test set. Specifically, for each subset, we randomly pick $60\%$ of the data from training, $20\%$ for validation and $20\%$ for testing.

The dataset contains $5,000$ frames acquired at $1\ fps$ for each scene-view pair, for a total of $30,000$ frames in the whole dataset. 
Each image has been collected at the resolution of $800 \times 600$ pixels. 
The time of the day and weather has been generated randomly. For each image, the simulator also produces a ground truth semantic segmentation map which associates each pixel of the scene to one of $13$ classes: \textit{buildings}, \textit{fences}, \textit{pedestrians}, \textit{poles}, \textit{road-lines}, \textit{roads}, \textit{sidewalks}, \textit{vegetation}, \textit{vehicles}, \textit{walls}, \textit{traffic signs}, \textit{other}, and \textit{unlabeled}.

This dataset allows to consider two types of source-target domain pairs: 1) ``point of view adaptation'' pairs, composed by two subsets from the same scene context but with different views (e.g., $A1-B1$, $A2-B2$, etc. in \figurename~\ref{fig:dataset}), 2) ``scene adaptation'' pairs, composed by two subsets belonging to different scene contexts (e.g., $A1-A2$, $A2-A3$, etc. in \figurename~\ref{fig:dataset}).

%% file: method.tex
\section{SceneAdapt Method}
\label{sec:method}
\begin{figure}[th!]
    \centering
        \includegraphics[width=1\linewidth]{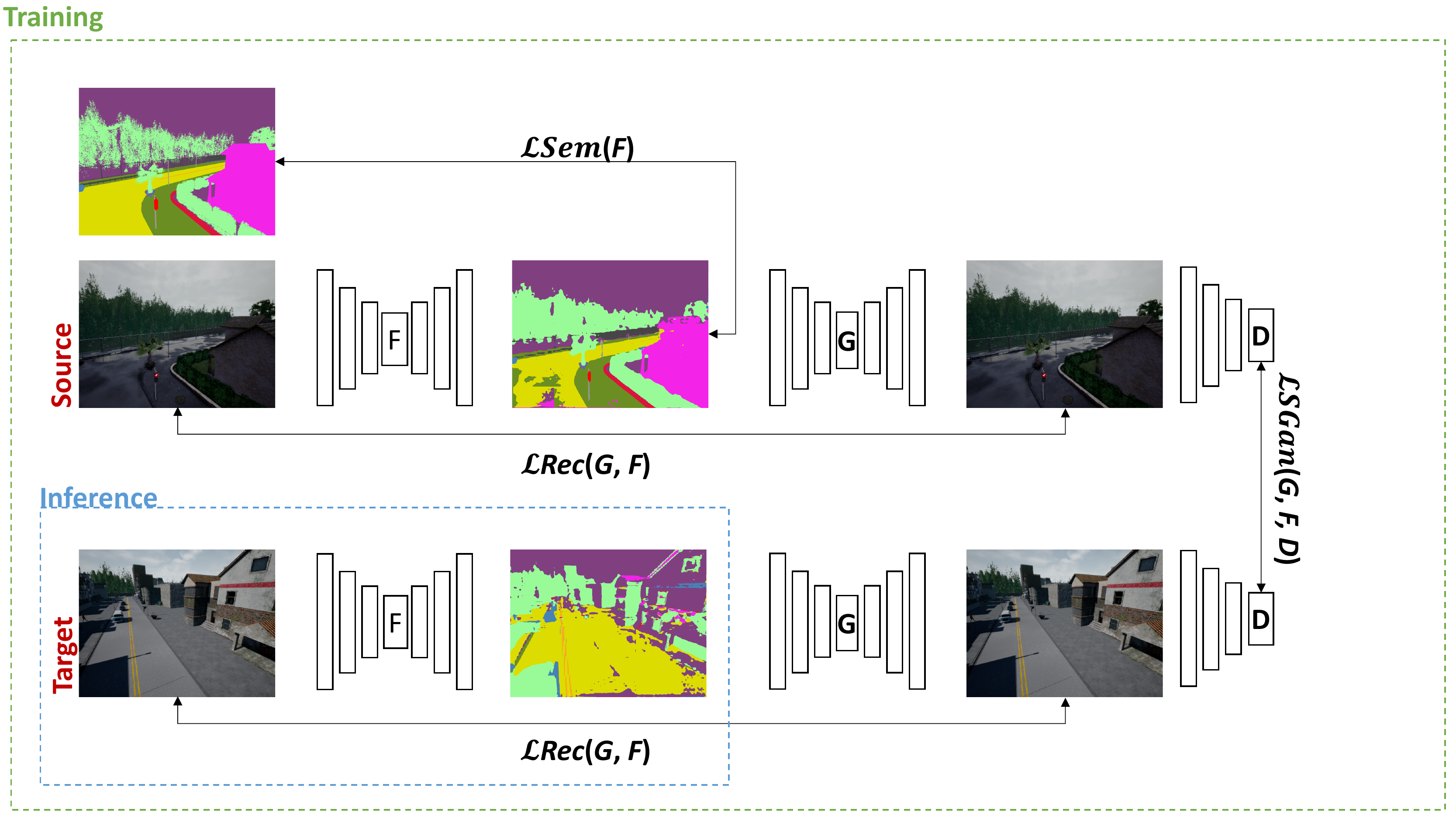}
    \caption{The proposed SceneAdapt architecture to train semantic segmentation networks for scene-based domain adaptation. During the training phase all networks are used employed (green ``Training'' dashed box), whereas only the part highlighted in blue is used for inference (blue ``Inference'' dashed box).}
\label{fig:model}
\vspace{-4mm}
\end{figure}
We propose SceneAdapt, a method to train a semantic segmentation network using labeled images from the source domain and only unlabeled images from the target domain. %
Indeed, at test time only the semantic segmentation module is needed to predict segmentation masks from images coming to both domains.

\subsection{Proposed Training Architecture}
\figurename~\ref{fig:model} illustrates the proposed training architecture, which consists of a the semantic network to be trained (\textit{F}), a generative network (\textit{G}), and a discriminator network (\textit{D}). The semantic network \textit{F} predicts semantic segmentation masks for images belonging to both the source and target domains. The ground truth segmentation masks of the source domain can be used to provide direct supervision to \textit{F} through a semantic segmentation loss. Direct supervision to \textit{F} cannot be provided for the target domain, where ground truth segmentation masks are not available. Therefore, we use the generator \textit{G} to reconstruct the input images from the generated ground segmentation masks and impose that the reconstructed images should be indistinguishable from the original ones. 
This encourages the semantic segmentation network \textit{F} to encode details useful for the reconstruction in the produced segmentation masks (e.g., small objects such as traffic lights and cars), which are generally lost in cross-domain testing. 
The reconstruction requirement is enforced using a reconstruction loss between original and generated images and an adversarial loss which makes use of a discriminator \textit{D} to improve the performance of the generator \textit{G} in an adversarial fashion. %

\textbf{Semantic Segmentation Component}
\textit{F} is responsible for predicting a segmentation mask for each input image, regardless of the related domain. 
In our experiments, we implement this module using the PSPNet architecture proposed by~\cite{zhao2017pyramid}, which is a state of art model. The network is implemented using a pre-trained ResNet backbone~\cite{he2016deep}. We would like to note that, in principle, any semantic segmentation network trainable end-to-end through gradient descent can be used to implement \textit{F}. 

\textbf{Generator Component}
\textit{G} is used to reconstruct the input images from the segmentation masks generated using \textit{F}. Specifically, the network \textit{G} takes over the pixel-wise class scores produced by the semantic segmentation network \textit{F} and generates an RGB image of the same dimensions as the input image. We implement the generator component using the architecture proposed by~\cite{johnson2016perceptual}, as implemented in~\cite{CycleGAN2017}.

\textbf{Discriminator Component}
\textit{D} allows to train the system using adversarial learning. Specifically, the discriminator is trained to distinguish between the input images and the generated ones, whereas the \textit{F} and \textit{G} components are joinlty trained to ``fool'' the discriminator, as proposed in~\cite{goodfellow2014generative}.
\textit{D} is implemented considering the PatchGANs architecture used in~\citep{isola2017image, li2016precomputed, ledig2016photo}. The network operates on $70 \times 70$ overlapping image patches. This patch-level discriminator architecture has fewer parameters than a full-image discriminator and can be applied to arbitrarily-sized images in a fully convolutional fashion as described in~\cite{isola2017image}.
\subsection{Loss Functions}
\label{loss}
The overall architecture illustrated in~\figurename~\ref{fig:model} is trained using a composite loss which encourages the network to produce good semantic segmentation masks for both domains. Specifically, the loss combines a semantic segmentation loss, a reconstruction loss and an adversarial loss. %

\textbf{Semantic Segmentation Loss\hspace{2mm}}
This loss is used to provide direct supervision to \textit{F}, when it is trained with images from the source domain. 
The loss is computed as the pixel-wise cross entropy between the inferred probability mask and the ground truth mask. 
Note that this loss can be applied only to image-label pairs of the source domain, since no ground truth segmentation masks are available for images of the target domain.
Specifically, let $x$ be an image from the source domain $S$, let $F(x)$ be the set of scores produced by the semantic segmentation network, and let $\mathcal{S}(F(x))$ be the softmax of $F(X)$, computed along the class scores independently for each pixel. 
Let $y$ be the ground truth segmentation mask associated to $x$, which indicates that a pixel $i$ belongs to class $y_{i}$. 
We define the semantic segmentation loss as follows:
\begin{align}
\mathcal{L}_{Sem}(F) = \mathbb{E}_{x \sim p_S(x)} [ - \sum_{i} log(\mathcal{S}(F(x))_{i,{y_i}})],
\end{align}
where $p_S$ denotes the distribution of the data of the source domain, $i$ iterates over the pixels of the training image $x$, $\mathcal{S}$ denotes the softmax function, and $\mathcal{S}(F(x))_{i,{y_i}}$ denotes the probability predicted for class $y_i$ and pixel $i$.

\textbf{Reconstruction Loss\hspace{2mm}}
As done in~\cite{CycleGAN2017}, we use an $L_1$ loss to enforce the correct reconstruction of the input images starting from the pixel-wise class scores produced by the semantic segmentation network (before Softmax). 
This loss is calculated for all the input-reconstructed image pairs of both the source and target domains. 
The reconstruction loss is defined as follows:
\begin{align}
 \mathcal{L}_{Rec}(G, F) = \mathbb{E}_{x \sim p_{ST}(x)} [||G(F(x)) - x||_1],
\end{align}
where $p_{ST}$ denotes the distribution of data belonging to both source and target domains, and $||\cdot||_1$ denotes the L1 norm. This loss is used to provide indirect supervision to the semantic network \textit{F} also for the unlabeled images belonging to the target domain, encouraging it to include in the segmentation mask domain-specific details useful for reconstruction.

\textbf{Adversarial Loss \hspace{2mm}} This loss allows to improve the performance of \textit{F} and \textit{G} in an adversarial fashion. Specifically, it forces the generator \textit{G} to produce better reconstructions and overcome the common limitations of regression-based losses for reconstruction purposes~\citep{mathieu2015deep}.
The loss is applied to images generated for both the source and target domain and is defined as follows:
\begin{align}
\mathcal{L}_{GAN}(G, F, D)=\mathbb{E}_{x \sim p_{ST}(x)} [log (D(x))] + \\ \mathbb{E}_{x \sim p_{ST}(x)} [log (1 - D(G(F(x))))].
\end{align}
\textbf{Overall Loss\hspace{2mm}} The overall loss used to optimize the whole training architecture is defined as the sum of the three losses discussed in previous sections:
\begin{align}
 \mathcal{L}(G, F, D) =\mathcal{L}_{Sem}(F) + \mathcal{L}_{Rec}(G, F)+ \mathcal{L}_{GAN}(G, F, D).
\end{align}

%% file: results.tex
\section{Experimental Settings and Results}
\label{sec:settings}
We compare the proposed method with respect to the following baselines and state of the art domain adaptation techniques:

\textbf{No Adaptation (NA)\hspace{2mm}} This baseline is obtained training PSPNet, on the source domain and testing it on the target domain, without adaptation;

\textbf{Fine Tuning (FT)\hspace{2mm}} This is a strong baseline, obtained performing the fine-tuning procedure, i.e., training PSPNet on the target domain using ground truth annotations. It should be noted that, since this strong baseline uses the ground truth annotations, it is reported for reference only.

\textbf{Geometric Warp (WARP)\hspace{2mm}} This baseline tries to model the domain gap through geometrical considerations. In particular, it assumes that an affine transformation $\mathcal{H}$ between the source and target scenes exists and is known. Hence, we test this baseline only in the case of point of view adaptation, where a geometrical correspondence between the scenes can be established.
At training time, all images from the source domain are warped using $\mathcal{H}$ in order to match the geometry of the target domain. 
The same transformation is applied to the ground truth segmentation masks with nearest neighbor interpolation to recover the class of missing points. 
A PSPNet model is hence trained on the warped images. 
The network is tested directly on the target domain images, where no warp operation is needed. 
It should be noted that, as the source images need to be mapped to a higher resolution in order to preserve the scale of the objects appearing in the scene, this baseline is computationally expensive to train.

\textbf{ASEGNET\hspace{2mm}} A domain adaptation approach proposed by~\cite{DBLP:journals/corr/abs-1802-10349}. We use the official implementation provided by the authors for the experiments;

\textbf{LSD\hspace{2mm}} A domain adaptation method proposed by~\cite{DBLP:journals/corr/abs-1711-06969}. This method is composed by 
an encoder with a bottleneck, a decoder, and a semantic decoder. The encoder maps the input image to a spatial embedding space. The first decoder reconstructs the image form the embedded input, whereas the semantic decoder infers the semantic segmentation mask from the embedded input. Note that the main difference between the proposed LSD and the proposed method, is that we do not consider an embedded space. In contrast, we enforce that the produced semantic segmentation masks are sufficient to reconstruct the input. This is equivalent to considering a highly semantic embedding space induced by the softmax predictions.  We use the official implementation provided by the authors for the experiments. 

We perform two sets of experiments to assess the performance of the proposed method in the contexts of point of view adaptation and general scene adaptation. 
In the first set of experiments (point of view adaptation), the source and target domains are related to images acquired in the same scene from different points of view. 
In the second set of experiments (general scene adaptation), source and target domains are related to images acquired in different, non-overlapping scenes.

All algorithms have been trained till convergence.
The weights of the best epoch (according to the validation set) are selected for the evaluation. 
For the baselines (i.e. NA and FT) and our method, we use PSPNet with Resnet-101 as backbone and we optimize using Stochastic Gradient Descent (SGD) with momentum equal to~$0.9$, initial learning rate equal to~$0.007$, weight decay equal to~$(1 - i/3750)^{0.9}$ (where~$i$ is the current iteration, and $3750$ is the total number of iterations) and batch size equal to~$8$.
The proposed method is optimized using the Adam optimizer, with initial learning rate equal to~$0.0002$ and batch size equal to~$1$.
LSD~\citep{DBLP:journals/corr/abs-1711-06969} is trained using SGD with momentum equal to $0.9$, initial learning equal to~$1.0 \cdot 10^{-5}$ and weight decay coefficient equal to $0.0005$.
ASEGNET~\citep{DBLP:journals/corr/abs-1802-10349} is trained using SGD with momentum $0.9$, initial learning equal to~$2.5 \cdot 10^{-4}$ and weight decay coefficient equal to $0.0005$. All the results have been evaluated using per-class accuracy ($c_{acc}$) and mean intersection over union ($m_{iou}$) as defined in~\cite{long2015fully}.

\subsection{Comparison with the state of the art}

\tablename~\ref{table:view} reports the point of view adaptation results\footnote{The results are averaged over the following source-target image pairs: $A1-B1$, $A2-B2$, $A3-B3$, $B1-A1$, $B2-A2$, $B3-A3$.}. Best per-row results are highlighted in bold numbers in the table. The cases in which the best method outperforms the \textit{FT} strong baseline are underlined.
The table reports the average results, as well as the breakdown according to the different classes.
\begin{table}[t]
	\tiny
	\centering
	\caption{Point of view adaptation results.}
	\label{table:view}
	\begin{tabular}{c|c|c|c|c||c}
	\multicolumn{6}{c}{\textbf{PER-CLASS ACCURACY}} \\ \hline
		& \textbf{NA} & \textbf{ASEGNET} & \textbf{LSD} & \textbf{SceneAdapt} & \textbf{\textit{FT}}   \\ \hline
		\textbf{Average} & 0.53 & 0.52 & 0.59 & \textbf{\underline{0.72}} & \textit{0.69}  \\ \hline
		\textbf{Unlabeled} & 0.27 & 0.80 & 0.74 & \textbf{0.90} & \textit{0.96}  \\
		\textbf{Buildings} & \textbf{0.96} & 0.93 & 0.53 & 0.95 & \textit{0.98}  \\
		\textbf{Fences} & 0.28 & 0.36 & \textbf{\underline{0.58}} & 0.55 & \textit{0.50}  \\
		\textbf{Other} & 0.29 & 0.32 & 0.56 & \textbf{\underline{0.60}} & \textit{0.46}  \\
		\textbf{Pedestrians} & 0.21 & 0.04 & 0.11 & \textbf{\underline{0.53}} & \textit{0.21}  \\
		\textbf{Poles} & 0.05 & 0.08 & \textbf{\underline{0.44}} & 0.25 & \textit{0.22}  \\
		\textbf{Road-lines} & 0.36 & 0.38 & \textbf{\underline{0.55}} & 0.48 & \textit{0.44}  \\
		\textbf{Roads} & \textbf{0.96} & 0.88 & 0.59 & 0.93 & \textit{0.98}  \\
		\textbf{Sidewalks} & \textbf{0.97} & 0.88 & 0.73 & 0.93 & \textit{0.98}  \\
		\textbf{Vegetation} & \textbf{\underline{0.89}} & 0.72 & 0.77 & 0.82 & \textit{0.85}  \\
		\textbf{Vehicles} & 0.66 & 0.48 & 0.45 & \textbf{\underline{0.94}} & \textit{0.84}  \\
		\textbf{Walls} & 0.75 & 0.83 & \textbf{\underline{0.88}} & 0.82 & \textit{0.88}  \\
		\textbf{T. Signs} & 0.20 & 0.13 & \textbf{\underline{0.78}} & 0.68 & \textit{0.73}  \\ \hline
		\multicolumn{6}{c}{\textbf{MEAN INTERSECTION OVER UNION}} \\ \hline
		\textbf{Average} & 0.32 & 0.37 & 0.34 & \textbf{0.57} & \textit{0.64}  \\ \hline
		\textbf{Unlabeled} & 0.26 & 0.75 & 0.48 & \textbf{0.85} & \textit{0.91}  \\
		\textbf{Buildings} & 0.74 & 0.87 & 0.49 & \textbf{0.93} & \textit{0.98}  \\
		\textbf{Fences} & 0.13 & 0.15 & 0.33 & \textbf{0.42} & \textit{0.42}  \\
		\textbf{Other} & 0.22 & 0.10 & 0.25 & \textbf{\underline{0.50}} & \textit{0.42}  \\
		\textbf{Pedestrians} & 0.09 & 0.03 & 0.09 & \textbf{0.13} & \textit{0.19}  \\
		\textbf{Poles} & 0.04 & 0.02 & 0.17 & \textbf{\underline{0.20}} & \textit{0.17}  \\
		\textbf{Road-lines} & 0.20 & 0.03 & 0.10 & \textbf{\underline{0.36}} & \textit{0.34}  \\
		\textbf{Roads} & 0.55 & 0.82 & 0.41 & \textbf{0.86} & \textit{0.95}  \\
		\textbf{Sidewalks} & 0.76 & 0.66 & 0.48 & \textbf{0.88} & \textit{0.92}  \\
		\textbf{Vegetation} & 0.52 & 0.49 & 0.64 & \textbf{0.71} & \textit{0.74}  \\
		\textbf{Vehicles} & 0.23 & \textbf{0.37} & 0.33 & 0.26 & \textit{0.78}  \\
		\textbf{Walls} & 0.28 & 0.52 & 0.31 & \textbf{0.71} & \textit{0.83}  \\
		\textbf{T. Signs} & 0.16 & 0.03 & 0.39 & \textbf{0.63} & \textit{0.70}  \\ \hline
	\end{tabular}
	\vspace{-4mm}
\end{table}
\begin{table}[t]
	\centering
	\caption{Results of the WARP baseline for the $A1$-$B1$ pair.} 		\tiny
	\label{table:affine}
	\begin{tabular}{ccc}
		& \textbf{Per Class Accuracy} & \textbf{MIoU}\\ \hline
		NA    & 0.47  & 0.43 \\
		WARP    & 0.58  & 0.46 \\
		SceneAdapt  & \underline{\textbf{0.75}} & \textbf{0.62} \\ \hline
		\textit{FT}  & \textit{0.72} & \textit{0.66} \\ \hline
	\end{tabular}
\end{table}
The proposed method outperforms with a good margin all competitors (in average) considering both per-class accuracy and mean intersection over union. Interestingly, while the strong baseline \textit{FT} reports in general the best results, the proposed method outperforms \textit{FT} for some classes such as ``Fences'', ``Pedestrians'', ``Vehicles'' and ``T. Signs''. 
This suggests that a) the proposed training architecture benefits from joint domain training, as compared to training only with data coming from one domain (as in FT) and b) the proposed method can effectively prevent over-fitting improving the segmentation of small objects. 
Similar gains can be observed for the mean intersection over union measure.
\tablename~\ref{table:affine} compares the performance of the WARP baseline with respect to \textit{NA}, \textit{FT} and the proposed method. Please note that, since the WARP baseline is very computationally expensive, we perform these experiments only on the \textit{A1-B1} pair. It is worth noting that beside being very computationally expensive to train, the WARP baseline only allows to obtain minor improvements \textit{NA}, reaching sub-optimal performance with respect to the proposed approach. This suggests that, while a geometric prior can be helpful, the proposed framework implicitly allows to learn some of the differences in the scene layouts arising from different points of view.

\begin{table}[t]
	\tiny
	\centering
	\caption{Scene adaptation results.}
	\label{table:episode}
	\begin{tabular}{c|c|c|c|c||c}
		\multicolumn{6}{c}{\textbf{PER-CLASS ACCURACY}} \\ \hline
		& \textbf{NA} & \textbf{ASEGNET} & \textbf{LSD} & \textbf{SceneAdapt} & \textbf{\textit{FT}}  \\ \hline
		
		\textbf{Average} & 0.27 & \textbf{0.37} & 0.32 & 0.34 & \textit{0.68}  \\ \hline
		\textbf{Unlabeled} & 0.53 & 0.54 & 0.39 & \textbf{\underline{0.69}} & \textit{0.23}  \\
		\textbf{Buildings} & 0.09 & 0.18 & 0.17 & \textbf{0.25} & \textit{0.99}  \\
		\textbf{Fences} & 0.07 & 0.03 & \textbf{0.25} & 0.01 & \textit{0.52}  \\
		\textbf{Other} & 0.02 & \textbf{0.07} & 0.02 & 0.00 & \textit{0.70}  \\
		\textbf{Pedestrians} & 0.05 & 0.01 & 0.03 & \textbf{0.13} & \textit{0.38}  \\
		\textbf{Poles} & 0.03 & 0.11 & \textbf{0.25} & 0.03 & \textit{0.25}  \\
		\textbf{Road-lines} & 0.13 & \textbf{\underline{0.66}} & 0.49 & 0.52 & \textit{0.43}  \\
		\textbf{Roads} & \textbf{0.84} & 0.78 & 0.75 & 0.77 & \textit{0.99}  \\
		\textbf{Sidewalks} & 0.67 & \textbf{0.69} & 0.67 & 0.50 & \textit{0.98}  \\
		\textbf{Vegetation} & 0.41 & 0.22 & 0.39 & \textbf{0.49} & \textit{0.94}  \\
		\textbf{Vehicles} & 0.61 & 0.57 & 0.35 & \textbf{0.81} & \textit{0.89}  \\
		\textbf{Walls} & \textbf{0.00} & \textbf{0.00} & \textbf{0.00} & \textbf{0.00} & \textit{0.94}  \\
		\textbf{T. Signs} & 0.04 & \textbf{\underline{0.99}} & 0.73 & 0.18 & \textit{0.65}  \\ \hline
		\multicolumn{6}{c}{\textbf{MEAN INTERSECTION OVER UNION}} \\ \hline
		\textbf{Average} & 0.19 & 0.18 & 0.20 & \textbf{0.22} & \textit{0.47}  \\ \hline
		\textbf{Unlabeled} & 0.30 & 0.46 & 0.17 & \textbf{\underline{0.50}} & \textit{0.22}  \\
		\textbf{Buildings} & 0.07 & 0.09 & 0.25 & \textbf{0.25} & \textit{ 0.90}  \\
		\textbf{Fences} & \textbf{0.06} & 0.01 & 0.02 & 0.01 & \textit{0.21}  \\
		\textbf{Other} & \textbf{0.02} & 0.01 & 0.01 & 0.00 & \textit{0.40}  \\
		\textbf{Pedestrians} & \textbf{0.03} & 0.01 & \textbf{0.03} & 0.02 & \textit{0.15}  \\
		\textbf{Poles} & 0.01 & 0.01 & \textbf{0.08} & 0.02 & \textit{0.25}  \\
		\textbf{Road-lines} & 0.13 & 0.07 & 0.27 & \textbf{\underline{0.40}} & \textit{0.33}  \\
		\textbf{Roads} & 0.62 & \textbf{0.71} & 0.65 & 0.59 & \textit{0.88}  \\
		\textbf{Sidewalks} & \textbf{0.45} & 0.44 & 0.40 & 0.43 & \textit{0.88}  \\
		\textbf{Vegetation} & 0.28 & 0.10 & 0.23 & \textbf{0.33} & \textit{0.54}  \\
		\textbf{Vehicles} & 0.40 & \textbf{0.48} & 0.25 & 0.21 & \textit{0.71}  \\
		\textbf{Walls} & \textbf{0.00} & \textbf{0.00} & \textbf{0.00} & \textbf{0.00} & \textit{0.71}  \\
		\textbf{T. Signs} & 0.04 & 0.00 & \textbf{\underline{0.31}} & 0.16 & \textit{0.07}  \\ \hline
	\end{tabular}
	\vspace{-4mm}
\end{table}

\tablename~\ref{table:episode} reports the scene adaptation results\footnote{These results are averaged over the following source-target image pairs: $A1-A2$, $A1-A3$, $A2-A1$, $A2-A3$, $A3-A1$, $A3-A2$.}.
The proposed method outperforms all competitors in the case of mean intersection over union and achieves state-of-the-art results in the case of per-class accuracy. The results highlight that general scene adaptation is 
much more difficult than point of view adaptation. Specifically, the proposed method tends to achieve sub-optimal results as compared to the \textit{FT} strong baseline. This suggests that, due to the radical change 
of the image layout due to the very different scenes, the proposed method is not always able to accurately segment small objects such as ``pedestrians'' and ``traffic lines'', albeit it generally improves over the NA baseline.
A closer comparison between the LSD method and the proposed approach suggests that using the embedding space induced by the softmax predictions as done by our method is beneficial in the view adaptation scenario for almost all classes with respect to the mean intersection over union measure. In the scene adaptation scenario, our embedding works better with larger classes such as ``vegetation''.

\begin{figure}[h!]
	\hspace{0.02\linewidth}
	\begin{minipage}{0.7\linewidth}
	
		\tiny
		\begin{tabular}{cc|c}
			& \multicolumn{2}{c}{\includegraphics[width=\textwidth]{images/legend}} \\
			
			\raisebox{4\normalbaselineskip}[0pt][0pt]{\rotatebox[origin=c]{90}{\textit{Source}}} &
			\includegraphics[width=0.49\textwidth]{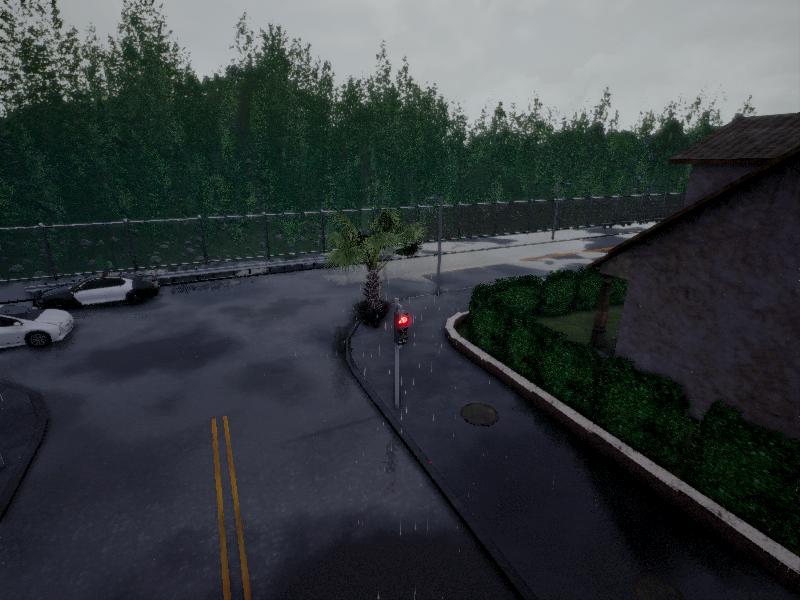} &
			
			\includegraphics[width=0.49\textwidth]{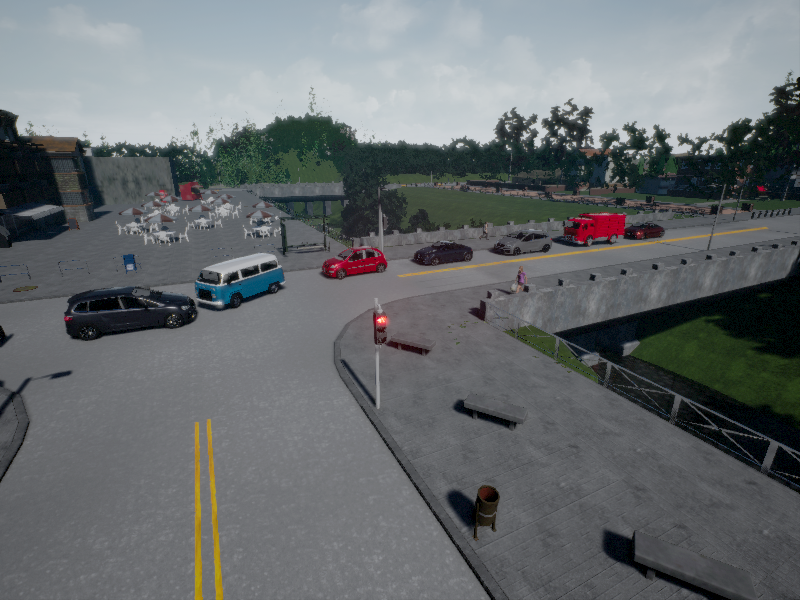} \\
			
			\raisebox{4\normalbaselineskip}[0pt][0pt]{\rotatebox[origin=c]{90}{\textit{Target}}} &
			\includegraphics[width=0.49\textwidth]{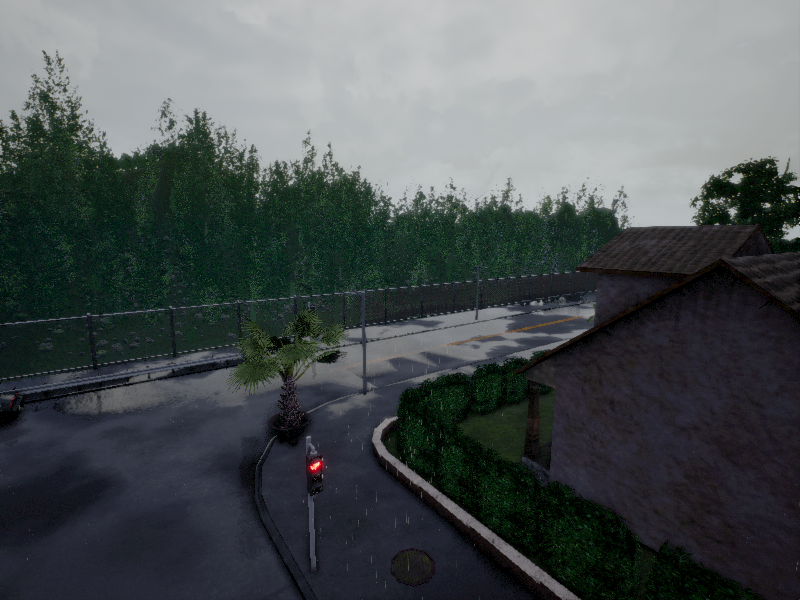} &
			
			\includegraphics[width=0.49\textwidth]{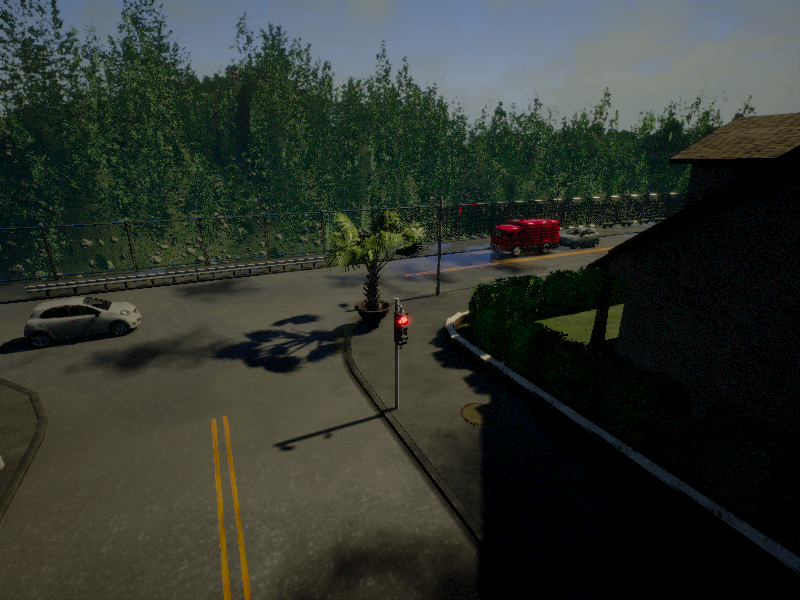}  \\
			
			\raisebox{4\normalbaselineskip}[0pt][0pt]{\rotatebox[origin=c]{90}{\textit{NA}}} &
			\includegraphics[width=0.49\textwidth]{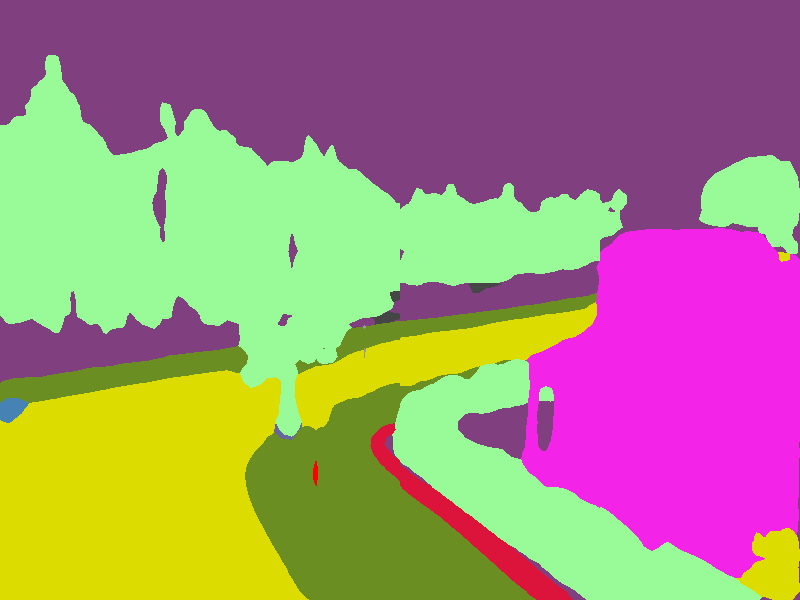} &
			
			\includegraphics[width=0.49\textwidth]{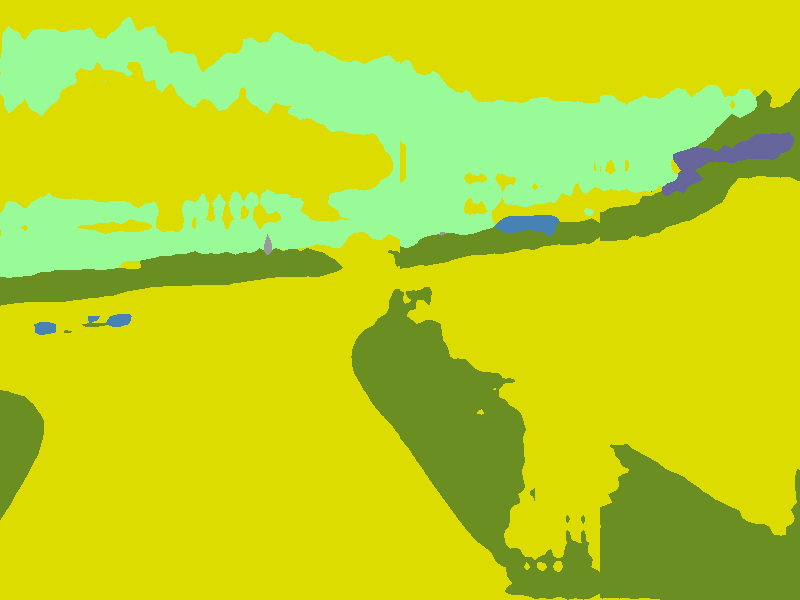} \\
			
			\raisebox{4\normalbaselineskip}[0pt][0pt]{\rotatebox[origin=c]{90}{\textit{ASEGNET}}} &
			\includegraphics[width=0.49\textwidth]{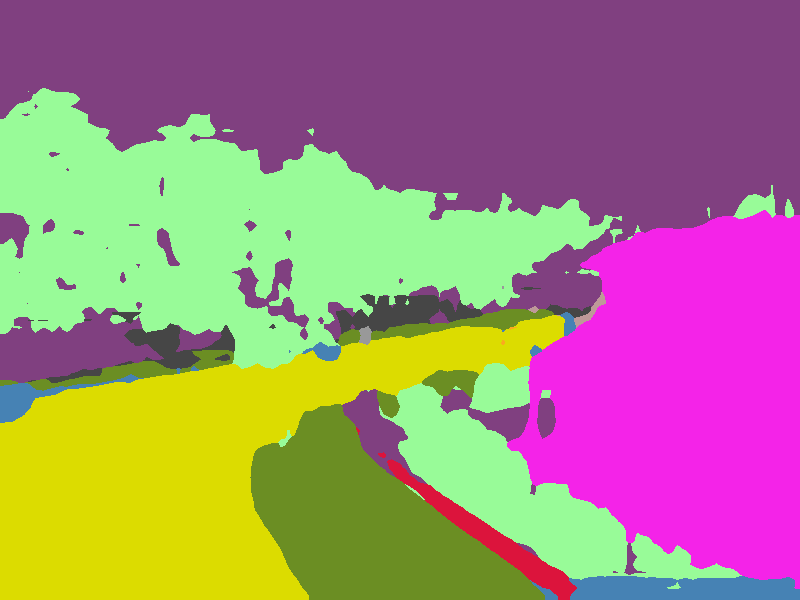}  &
			
			\includegraphics[width=0.49\textwidth]{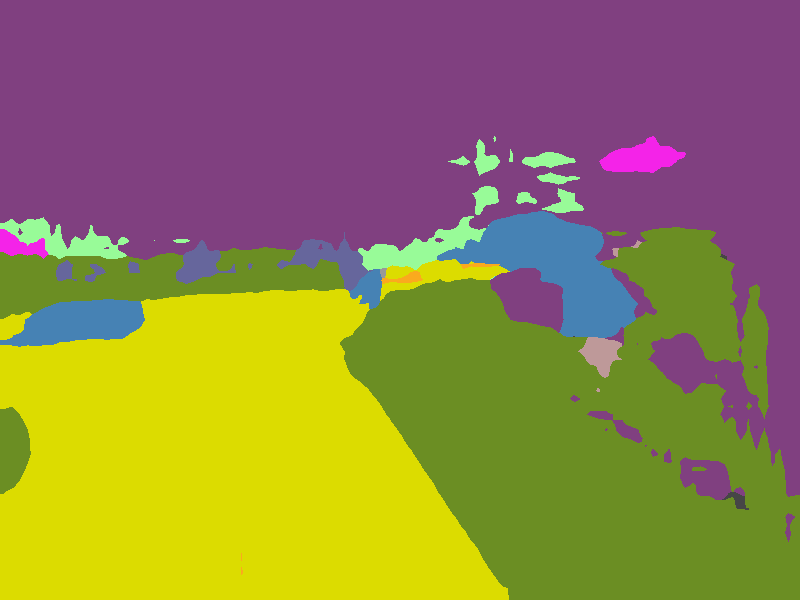} \\
			
			\raisebox{4\normalbaselineskip}[0pt][0pt]{\rotatebox[origin=c]{90}{\textit{LSD}}} &
			\includegraphics[width=0.49\textwidth]{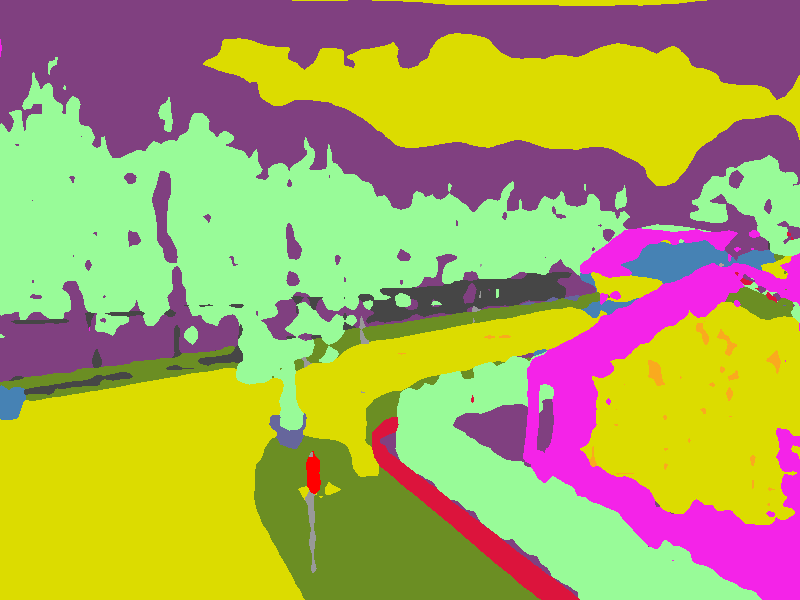} &
			
			\includegraphics[width=0.49\textwidth]{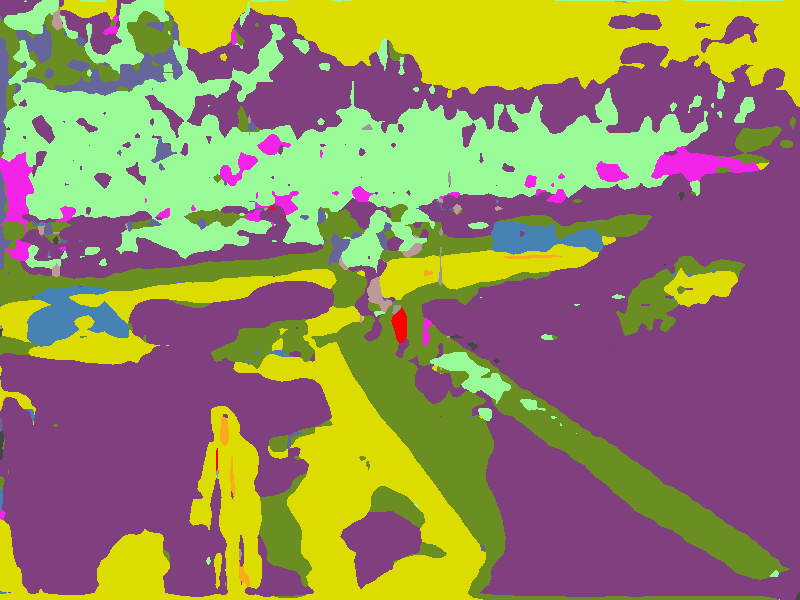} \\
			
			\raisebox{4\normalbaselineskip}[0pt][0pt]{\rotatebox[origin=c]{90}{\textit{SceneAdapt}}} &
			\includegraphics[width=0.49\textwidth]{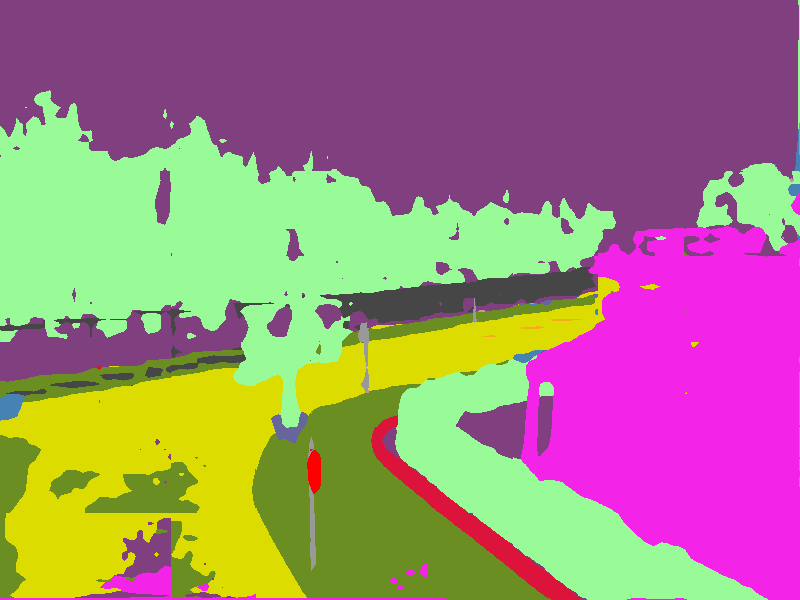} &
			
			\includegraphics[width=0.49\textwidth]{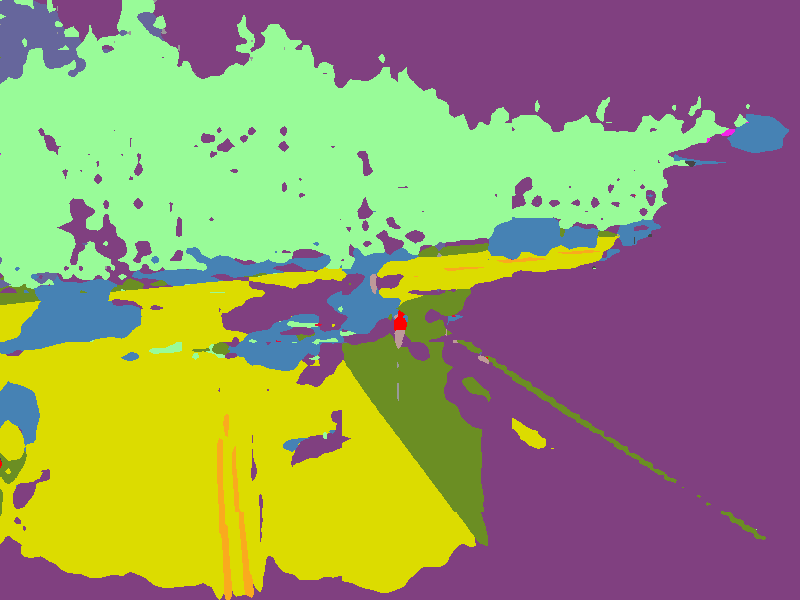} \\
			
			\raisebox{4\normalbaselineskip}[0pt][0pt]{\rotatebox[origin=c]{90}{\textit{FT}}} &
			\includegraphics[width=0.49\textwidth]{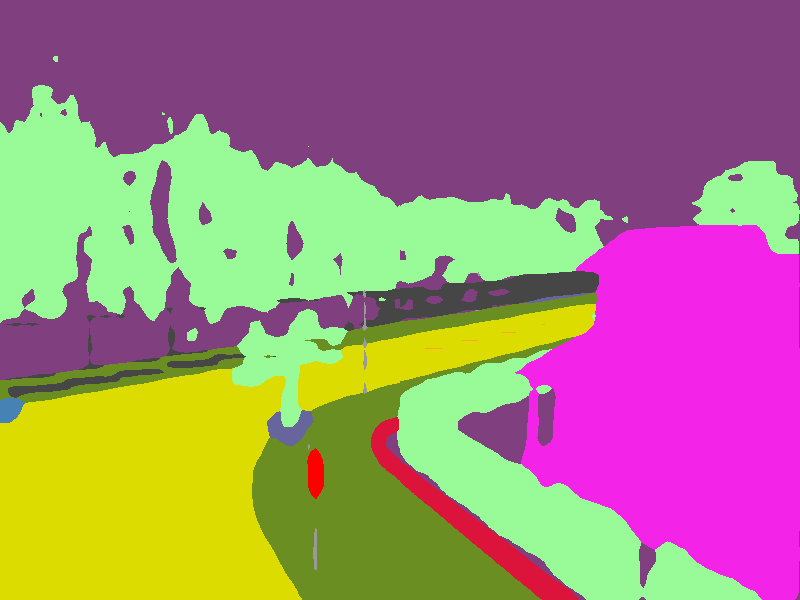} &
			
			\includegraphics[width=0.49\textwidth]{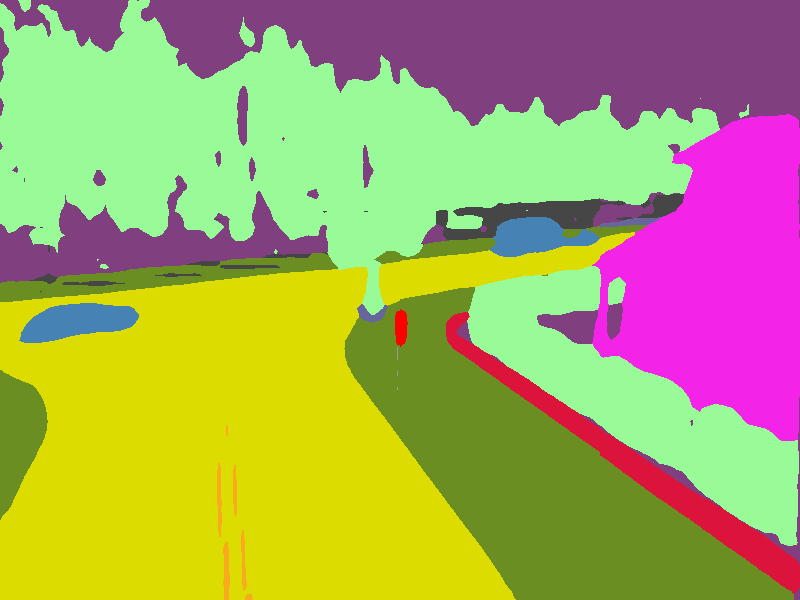} \\
			
			\raisebox{4\normalbaselineskip}[0pt][0pt]{\rotatebox[origin=c]{90}{\textit{GT}}} &
			\includegraphics[width=0.49\textwidth]{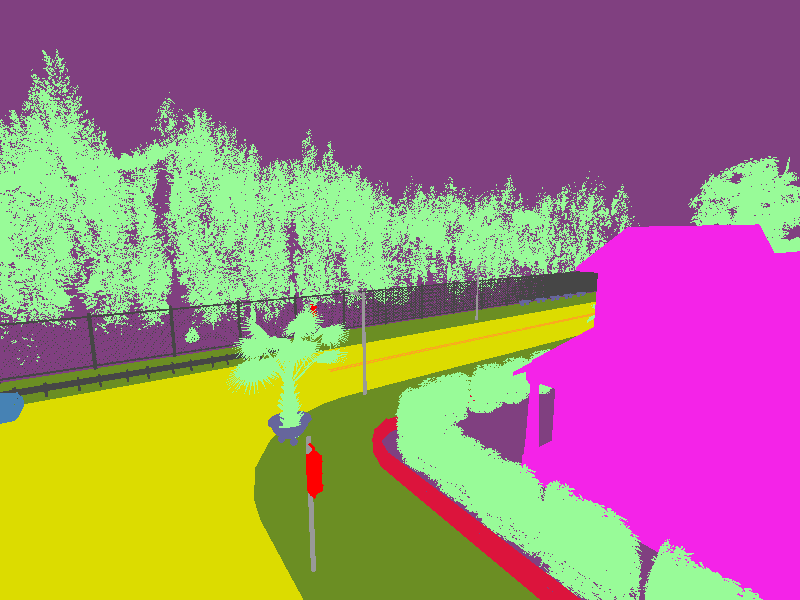} &
			
			\includegraphics[width=0.49\textwidth]{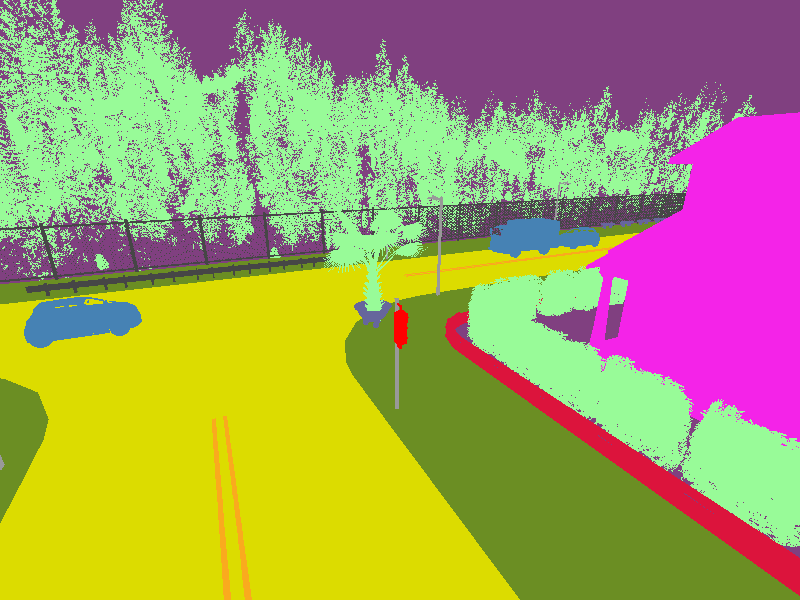} \\
			& \multicolumn{1}{c|}{\normalsize (a)} & \multicolumn{1}{c}{\normalsize (b)}
		\end{tabular}
	\end{minipage}
	\caption{Qualitative comparisons of the considered method on the view adaptation (a) and scene adaptation (b) tasks.}
	\label{figure:results:view}
	\vspace{-5mm}
\end{figure}
\figurename~\ref{figure:results:view} reports some qualitative examples of the compared methods on the tasks of view and scene adaptation.

\begin{table*}[th!]
	\scriptsize
	\centering
	\caption{Results of the ablation study.}
	\label{table:ablation}
	\begin{tabular}{cccc}
		\textbf{Loss} & \textbf{Adaptation} & $\mathbf{c_{acc}}$ & $\mathbf{m_{iou}}$\\ \hline
		$\mathcal{L}_{SemS}(F) + \mathcal{L}_{Rec}(G, F)$     & point of view       & \textbf{0.77}  & \textbf{0.67} \\
		$\mathcal{L}_{SemS}(F) + \mathcal{L}_{GAN}(G, F, D)$  & point of view       & 0.71  & 0.55 \\
		$\mathcal{L}_{SemS}(F) + \mathcal{L}_{Rec}(G, F) + \mathcal{L}_{GAN}(G, F, D)$  & point of view  & 0.75 & 0.62 \\ \hline
		$\mathcal{L}_{SemS}(F) + \mathcal{L}_{Rec}(G, F)$     & scene    & 0.33  & 0.20 \\
		$\mathcal{L}_{SemS}(F) + \mathcal{L}_{GAN}(G, F, D)$  & scene    & 0.38  & \textbf{0.24} \\
		$\mathcal{L}_{SemS}(F) + \mathcal{L}_{Rec}(G, F) + \mathcal{L}_{GAN}(G, F, D)$ & scene & \textbf{0.39} & \textbf{0.24} \\ \hline
	\end{tabular}
	\vspace{-4mm}
\end{table*}

\subsection{Ablation study}
To assess the contribution of the reconstruction and adversarial losses described in Section~\ref{loss}, 
we tested three versions of the proposed method on the $A1-B1$ pair for view adaptation and on the $A1-A2$ pair for scene adaptation. 
Each considered version of the proposed method is trained using different combinations of loss functions as it is shown in \tablename~\ref{table:ablation}.
From the results, it can be observed that dropping the adversarial term in the loss function allows to obtain a boost of $+2\%$ in per-class accuracy and $+5\%$ 
in mean intersection over union, whereas dropping the reconstruction term yields worse results. Conversely, for general scene adaptation, the L1 reconstruction term has 
marginal importance, while combining both reconstruction and adversarial losses leads to the best results. 
These observations suggest that the L1 term can impose a strong geometric prior, which is useful in the case of point of view adaptation, where the scenes present similar, yet distinct, 
layouts. When geometric layouts are more diverse, as in the case of scene adaptation, the adversarial term is more beneficial and can overcome the 
``regression-to-the-mean'' effect which could be induced by the L1 loss.

We also assessed if the proposed method allows to improve segmentation results on the source domain thanks to the additional unlabeled images coming from the target domain seen during the training phase.
To assess generalization in the case of point of view adaptation, we trained the proposed method on the training sets of the source-target pair $A1-B1$ and tested it on the test set of $A1$ (the source domain). 
\tablename~\ref{table:source_domain} compares the results with respect to the NA baseline. As can be noted, the proposed method allows to improve performance on the source domain in all cases. 
This suggests a regularizing effect induced by the use of unlabeled images from the target domain. \tablename~\ref{table:source_domain_2} further compares the proposed SceneAdapt method with respect to the NA baseline reporting per-class measures. Interestingly, most of the improvement is obtained with respect to classes denoting small scene elements, such as ``fences'' (from $0.4/0.31$ to $0.63/0.44$ - Per-class Acc./Mean IoU) and ``pedestrians'' (from $0.16/0.15$ to $0.41/0.27$), which are the most sensitive to the changes of points of view. This suggests that the proposed approach successfully encourages the semantic segmentation network to focus on task-relevant details.

\begin{table*}[th!]
	\setlength{\tabcolsep}{0.2em}
	\scriptsize
	\centering
	\caption{Improvement on the source domain.}
	\label{table:source_domain}
	\begin{tabular}{cccccc}
		\textbf{Source} & \textbf{Target} & \textbf{Adaptation} & \textbf{Method} &  \textbf{Measure} & \textbf{Test Results} \\
		\hline
		$A1$ & -    & -        & No Adaptation & Per Class Accuracy & $0.70$ \\
		$A1$ & -    & -        & No Adaptation & Mean IoU & $0.64$ \\ \hline
		$A1$ & $B1$ & point of view & SceneAdapt & Per Class Accuracy & $0.74$ \\
		$A1$ & $B1$ & point of view & SceneAdapt & Mean IoU & $0.65$ \\ \hline
		$A1$ & $A2$ & scene & SceneAdapt & Per Class Accuracy & $0.75$ \\
		$A1$ & $A2$ & scene & SceneAdapt & Mean IoU & $0.66$ \\  \hline
	\end{tabular}
	\vspace{-4mm}
\end{table*}

\begin{table}[t]
	\tiny
	\centering
	\caption{Results on the source domain for point of view adaptation.}
	\label{table:source_domain_2}
	\begin{tabular}{c|c|c|c|c}
	    & \multicolumn{2}{c|}{\textbf{PER-CLASS ACC.}} & \multicolumn{2}{c}{\textbf{MEAN IoU}} \\ \hline
		& \textbf{NA} & \textbf{SceneAdapt} & \textbf{NA} & \textbf{SceneAdapt} \\ \hline
		\textbf{Average}    & 0.69 & \textbf{0.76} & \textbf{0.64} & \textbf{0.64} \\ \hline
		\textbf{Unlabeled}  & 0.77 & \textbf{0.79} & \textbf{0.70} & \textbf{0.70} \\
		\textbf{Buildings}  & \textbf{0.98} & 0.97 & \textbf{0.97} & 0.84 \\
		\textbf{Fences}     & 0.40 & \textbf{0.63} & 0.31 & \textbf{0.44} \\
		\textbf{Other}      & 0.65 & \textbf{0.83} & 0.50 & \textbf{0.68} \\
		\textbf{Pedestrians}& 0.16 & \textbf{0.41} & 0.15 & \textbf{0.27} \\
		\textbf{Poles}      & 0.15 & \textbf{0.45} & 0.14 & \textbf{0.32} \\
		\textbf{Road-lines} & 0.43 & \textbf{0.67} & 0.39 & \textbf{0.52} \\
		\textbf{Roads}      & \textbf{0.99} & 0.92 & \textbf{0.97} & 0.88 \\
		\textbf{Sidewalks}  & \textbf{0.97} & 0.85 & \textbf{0.93} & 0.73 \\
		\textbf{Vegetation} & \textbf{0.94} & 0.90 & \textbf{0.78} & 0.77 \\
		\textbf{Vehicles}   & \textbf{0.88} & 0.83  & \textbf{0.84} & 0.73 \\
		\textbf{Walls}      & \textbf{0.96} & 0.73 & \textbf{0.90} & 0.69 \\
		\textbf{T. Signs}   & 0.76 & \textbf{0.93}  & \textbf{0.74} & 0.76 \\ \hline
	\end{tabular}
	\vspace{-4mm}
\end{table}